%% file: manuscript.tex
\documentclass[10pt,twocolumn,letterpaper]{article}

\usepackage{cvpr}              

\usepackage{graphicx}
\usepackage{amsmath}
\usepackage{amssymb}
\usepackage{booktabs}

\newcommand{\cmark}{\ding{51}}%
\newcommand{\xmark}{\ding{55}}%

\usepackage{algorithm}
\usepackage{algpseudocode}

\usepackage{multirow}
\usepackage{color}

\usepackage{booktabs,colortbl,tabularx}
\usepackage{pifont}%

\def\ie{{\it i.e.}}

\usepackage{newfloat}
\usepackage{listings}
\floatstyle{ruled}
\newfloat{listing}{tb}{lst}{}
\floatname{listing}{Listing}

\usepackage[pagebackref,breaklinks,colorlinks]{hyperref}

\usepackage[capitalize]{cleveref}
\crefname{section}{Sec.}{Secs.}
\Crefname{section}{Section}{Sections}
\Crefname{table}{Table}{Tables}
\crefname{table}{Tab.}{Tabs.}

\begin{document}

\title{Multi-Scale Self-Contrastive Learning with Hard Negative Mining for Weakly-Supervised Query-based Video Grounding}

\input{author}

\maketitle

\input{SECTIONS/00_Abstract/manuscript}

\input{SECTIONS/10_Introduction/manuscript}

\input{SECTIONS/20_Related_Work/manuscript}

\input{SECTIONS/30_Methodology/manuscript}

\input{SECTIONS/40_Experiments/manuscript}

\input{SECTIONS/50_Ablation_Study/manuscript}

\input{SECTIONS/60_Conclusion/manuscript}

{\small
\bibliographystyle{ieee_fullname}
\bibliography{reference}
}

\end{document}

%% file: author.tex
\author{Shentong Mo \\
Carnegie Mellon University\\
Pittsburgh, United States\\
{\tt\small shentonm@andrew.cmu.edu}
\and
Daizong Liu\\
Peking University\\
Beijing, China\\
{\tt\small dzliu@stu.pku.edu.cn}
\and
Wei Hu\thanks{Corresponding author.}\\
Peking University\\
Beijing, China\\
{\tt\small  forhuwei@pku.edu.cn}
}

%% file: SECTIONS/00_Abstract/manuscript.tex
\begin{abstract}

Query-based video grounding is an important yet challenging task in video understanding, which aims to localize the target segment in an untrimmed video according to a sentence query. 
Most previous works achieve significant progress by addressing this task in a fully-supervised manner with segment-level labels, which require high labeling cost.  
Although some recent efforts develop weakly-supervised methods that only need the video-level knowledge, they generally match multiple pre-defined segment proposals with query and select the best one, which lacks fine-grained frame-level details for distinguishing frames with high repeatability and similarity within the entire video.
To alleviate the above limitations, we propose a self-contrastive learning framework to address the query-based video grounding task under a weakly-supervised setting. 
Firstly, instead of utilizing redundant segment proposals, we propose a new grounding scheme that learns frame-wise matching scores referring to the query semantic to predict the possible foreground frames by only using the video-level annotations. 
Secondly, since some predicted frames (\ie, boundary frames) are relatively coarse and exhibit similar appearance to their adjacent frames, we propose a coarse-to-fine contrastive learning paradigm to learn more discriminative frame-wise representations for distinguishing the false positive frames. 
In particular, we iteratively explore multi-scale hard negative samples that are close to positive samples in the representation space for distinguishing fine-grained frame-wise details, thus enforcing more accurate segment grounding.
Extensive experiments on two challenging benchmarks demonstrate the superiority of our proposed method compared with the state-of-the-art methods.




\end{abstract}

%% file: SECTIONS/10_Introduction/manuscript.tex

\section{Introduction}
\begin{figure*}
	    \centering
	\includegraphics[width=0.95\linewidth]{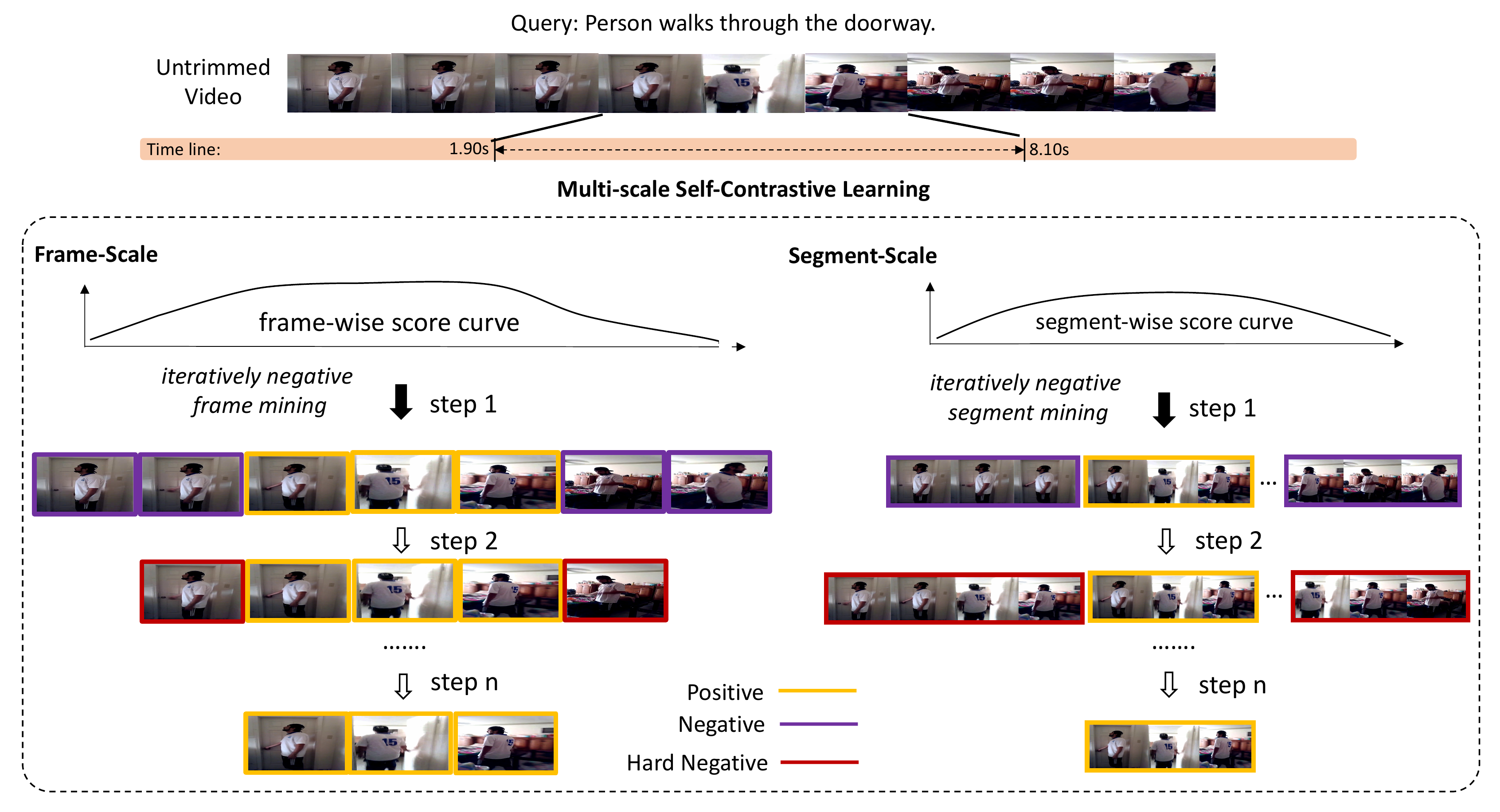}
   \caption{Illustration of the proposed multi-scale self-contrastive learning for weakly-supervised query-based video grounding. }
	\label{fig: title_img}
\end{figure*}

Query-based video grounding has attracted increasing attention due to its wide spectrum of applications in video understanding \cite{cheng2014temporal,shou2016temporal,liu2020violin,jiang2018recurrent}. 
This task aims to determine the start and end timestamps of a target segment in an untrimmed video that contains an activity semantically corresponding to a given sentence description, as shown in Figure~\ref{fig: title_img}. 
Most previous works \cite{Alpher06,Alpher07,wang2020temporally,yuan2020semantic,liu2021context,liu2020jointly} have achieved significant performance by addressing query-based video grounding in a fully-supervised manner, which however requires a large amount of segment-level annotations (location of the target segment in the video according to the semantic of the matched query). 
Such manual annotation is quite labor-intensive and time-consuming, thus limiting the wide applicability of query-based video grounding.

Recently, some weakly-supervised works \cite{mithun2019weakly,gao2019wslln,lin2020weakly,zhang2020counterfactual,ma2020vlanet} have been proposed to alleviate the above issue by only leveraging the video-level knowledge of matched video-query pairs without detailed segment labels. 
These methods generally pre-define multiple segment proposals, and employ video-level annotations as supervision to learn the segment-query matching scores for selecting the best one. 
However, the generated segment proposals are redundant and contain many negative (\ie, false) samples, resulting in inferior effectiveness and efficiency of the models. 
Further, as for the positive (\ie, correct) proposals covering the accurate foreground frames, they are of high similarity \cite{zeng2021multi} and require more sophisticated intra-modal recognition capabilities to distinguish. 
Especially for the boundary frames which exhibit similar visual appearance to the foreground frames in a certain segment, some of them are background frames that are hard to recognize. 
Once such segment is selected as the best one, the grounding performance will be degenerated due to the background noise.

To this end, we propose a novel Multi-scale Self-Contrastive Learning (MSCL) paradigm with hard negative mining strategy for weakly-supervised query-based video grounding, aiming to learn fine-grained frame-wise semantic matching by progressively sampling harder negative-positive frames for discriminative feature learning.  
In particular, instead of relying on redundant segment proposals for matching and selection, we propose to learn more fine-grained {\it frame-wise matching scores} to predict whether each frame is the foreground frame. 
Once the scores of successive frames are larger than a learnable threshold, they are taken to construct the predicted segment, leading to more efficient grounding. 
The threshold is acquired from the frame-wise scores learned and enhanced by our developed multi-scale self-contrastive learning paradigm.
We achieve such fine-grained frame-wise representation learning by the following twofold novelties.

{\bf Firstly}, since we only resort to video-level annotations with no access to the frame-level knowledge, we propose {\it frame-wise matching score prediction} to estimate the score of each frame matched to the video-level annotations in a weakly-supervised manner, as well as the frame-wise matching weight via an attention mechanism in order to choose possible foreground frames with query semantics.
{\bf Secondly}, in order to improve the frame-wise score prediction by enhancing the frame-wise representations, we propose a self-contrastive learning with multi-scale hard negative mining strategy, which especially discriminates frames adjacent to the ground truth target segment---referred to as {\it hard negative samples} as shown in Figure~\ref{fig: title_img}.
In particular, we dynamically set a range according to the previously estimated scores so as to select positive and negative samples, where samples with scores within the range serve as negative frames and those above the range as positive ones.  
The range is progressively updated to exploit harder negative samples that are more similar to the positive ones, which leads to more discriminative features learning.
Note that, our contrastive learning strategy is quite different from the previous vanilla ones \cite{nan2021interventional,zhang2020counterfactual,zhang2021video} in this task, since they all utilize a one-step algorithm to define the constant negative samples with coarse frame-level representations. Compared to them, we employ a multi-step process to iteratively mine the negative samples in a coarse-to-fine manner, defining harder negative samples and thus leading to more discriminative frame-wise representation learning.  
Further, we explore hard negative samples from different hierarchies---local frame-scale and nonlocal segment-scale, thus learning multi-scale intrinsic features.



Specifically, given the input video and query, we first encode both visual and textual features, and align their semantics by a video-query attention mechanism to learn the cross-modal interactions.
Next, we predict frame-wise matching scores referring to the interacted features for foreground frame (inside the target segment) localization, as well as the frame-wise contribution weights for their importance estimation, which are aggregated to compute the overall semantic dependency between the video-query pair. 
Further, we enhance frame-wise representations by learning more discriminative features via the proposed multi-scale self-contrastive learning strategy, where 
both frame-scale and segment-scale hard negative sampling are deployed in a coarse-to-fine manner. 

Our main contributions are summarized as follows:
\begin{itemize}
    \item We propose a novel self-contrastive learning framework for weakly-supervised query-based video grounding, which predicts fine-grained frame-wise matching scores referring to the query semantics for more accurate segment localization. 
    
    \item We propose a multi-scale hard negative mining in the self-contrastive learning to learn discriminative frame-wise representations by adaptively sampling hard negatives in the frame-scale and the segment-scale respectively, which captures both local and nonlocal intrinsic patterns.    
    
    \item Extensive experiments demonstrate that the proposed MSCL model outperforms the state-of-the-art method significantly over two challenging benchmarks.
\end{itemize}





%% file: SECTIONS/20_Related_Work/manuscript.tex
\section{Related Work}
\noindent \textbf{Fully-supervised query-based video grounding.}
Most existing works address the video grounding task in a fully-supervised manner, where both the annotations of video-sentence pairs and corresponding segment boundaries are given.
Traditional methods \cite{Alpher06,Alpher07} utilize the proposal-based framework that samples video segment proposals through dense sliding windows and subsequently integrate query with these proposal representations via a matrix operation.
To further mine the cross-modal interaction more effectively, some works \cite{wang2020temporally,yuan2020semantic,zhang2019man,zhang2019cross,Alpher10,Alpher33,liu2020reasoning,liu2021adaptive,liu2021progressively,liu2022exploring,liuExploring,liu2022unsupervised,liu2022memory} integrate the sentence representation with those pre-defined segment proposals individually, and then evaluate their matching relationships. The proposal with the highest matching score is selected as the target segment. Although the proposal-based methods can achieve significant performance, they severely rely on the quality of the proposals and are very time-consuming.
Instead of utilizing the segment proposals,
recent proposal-free methods \cite{Alpher08,Alpher11,Alpher12,chen2020rethinking} directly regress the temporal locations of the target segment. Specifically, they either regress the start/end timestamps based on the entire video representation \cite{Alpher08,Alpher11}, or predict at each frame to determine whether this frame is a start or end boundary \cite{Alpher12,chen2020rethinking}. These works are much more efficient than the proposal-based ones, but achieve relatively lower performance. However, both of proposal-based and proposal-free methods heavily rely on a large amount of human annotations that are hard to collect in practice.

\noindent \textbf{Weakly-supervised query-based video grounding.}
As manually annotating temporal boundaries of target moments is time-consuming, recent research attentions have been shifted to developing weakly-supervised video grounding models \cite{mithun2019weakly,chen2020look,song2020weakly,tan2021logan}, which only require video-level annotations. \cite{mithun2019weakly} proposed the first weakly-supervised model to learn a joint embedding space for video and query representations. \cite{gao2019wslln} develop a two-stream structure to measure the moment-query consistency and conduct moment selection simultaneously.
Although above methods have achieved promising performance, they are two-stage approaches that utilize multi-scale sliding windows to generate moment candidates, therefore suffering from inferior effectiveness and efficiency.
To address this issue, \cite{lin2020weakly,zhang2020counterfactual,ma2020vlanet} further improve the segment-sentence matching accuracy, and score all the moments sampled at different scales in a single pass.
\cite{wu2020reinforcement} employ a reinforcement learning framework to refine the segment boundary.
However, almost all of the existing methods rely on the segment proposals for matching and selection, which fail to capture and distinguish more fine-grained details among visually similar frames for acquiring more accurate segment boundaries.

\noindent \textbf{Contrastive Learning.}
Contrastive learning~\cite{he2019moco,chen2020simple} is a self-supervised learning paradigm that has demonstrated its effectiveness in many tasks, such as image classification, object detection and point cloud classification.
Previous works~\cite{nan2021interventional,zhang2020counterfactual,zhang2021video} also showed promising results of contrastive learning in video grounding.
Typically, \cite{nan2021interventional} proposed a dual contrastive learning loss function by utilizing video-level samples for video-to-video and video-to-query representation learning.
A Counterfactual Contrastive Learning framework~\cite{zhang2020counterfactual} is designed to distinguish video-level embeddings between counterfactual positive and negative samples for hard negative sampling.
Our model differs significantly from these methods:
1) We leverage the input single video only to perform contrastive learning over frames and segments, without resorting to different instances of videos as in previous works.
2) We propose multi-scale hard negative sampling at the frame scale and the segment scale to iteratively capture both local and non-local intrinsic feature representations.

%% file: SECTIONS/30_Methodology/manuscript.tex
\section{Methodology}

\begin{figure*}[!htb]
    \centering
    \includegraphics[width=0.95\textwidth]{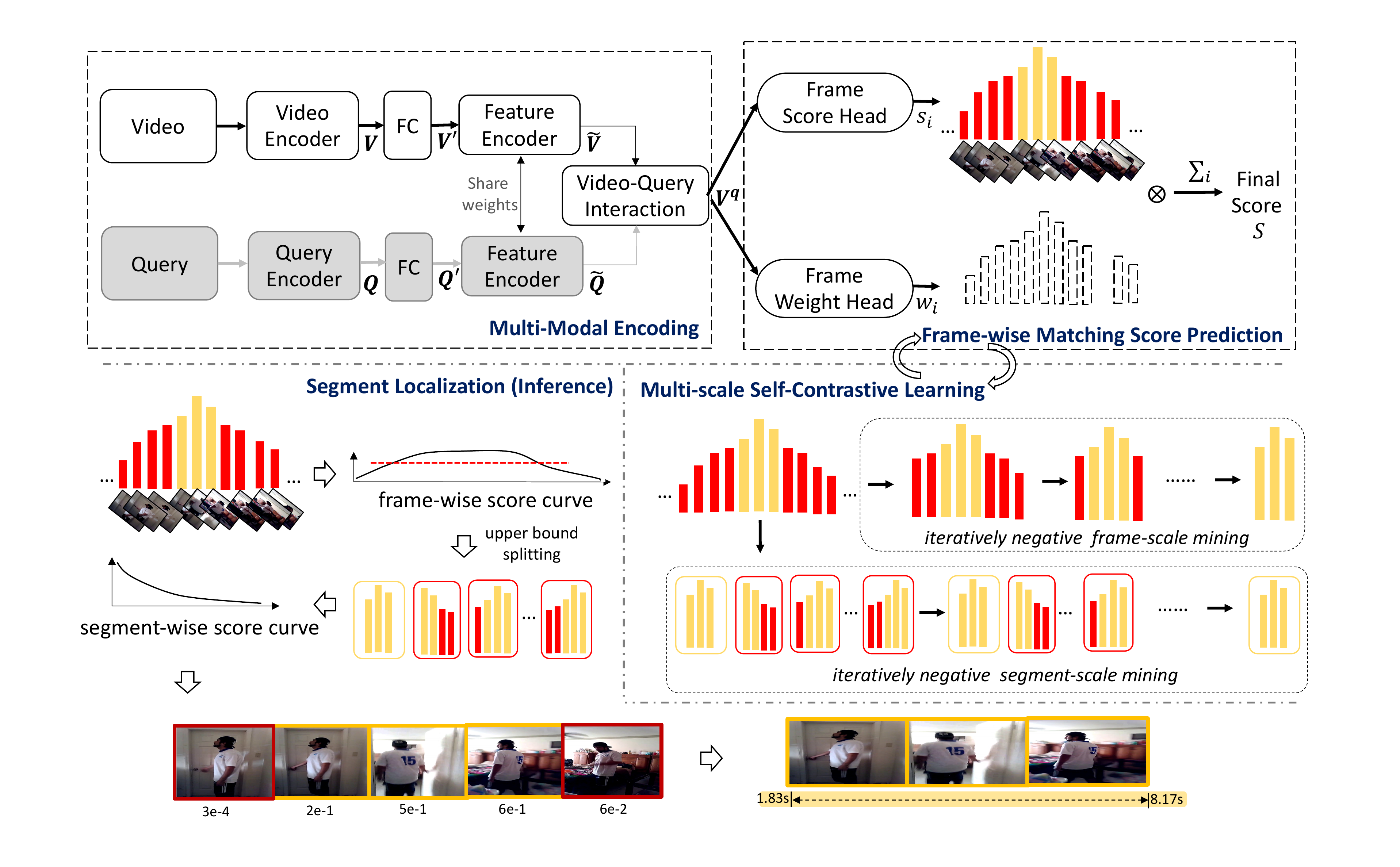}
    \caption{The overall framework of our proposed MSCL model. 
   }
    \label{fig:overview}
\end{figure*}

\subsection{Overview}
We focus on weakly-supervised query-based video grounding.
Given an untrimmed video and the language query, 
the goal is to localize the start and end time of the temporal moment corresponding to the language query.
As illustrated in Figure~\ref{fig:overview}, the proposed MSCL model mainly consists of four modules:

\begin{itemize}
    \item {\bf Multi-modal encoding.}
    Given the multi-modal input, we first employ video and query encoders to encode both visual and textual features, and then interact the cross-modal information for semantic alignment. 
    \item {\bf Frame-wise matching score prediction.}
    After generating query-specific video representations by the cross-modal interaction,
    we predict frame-wise matching scores and frame-wise matching weights referring to the aligned multi-modal features for choosing the possible foreground frames within the video.
    \item {\bf Multi-scale self-contrastive learning.}
    In order to improve the frame-wise score prediction by enhancing the frame-wise representations,
    we perform self-contrastive learning at both the frame-scale and the segment-scale with progressive hard negative mining to distinguish more fine-grained frame-wise details, thus enforcing more accurate segment grounding.
    
    \item {\bf Segment localization.}
    At the inference time, we first construct possible segments by choosing the consecutive frames with scores higher than the threshold, and then select the best segment by comparing the average scores of the internal frames within each segment.
\end{itemize}

We elaborate on the four modules in order as follows.

\subsection{Multi-Modal Encoding}
\noindent \textbf{Video and query encoders.}
For each video, following previous works, we first employ a pre-trained C3D model~\cite{du2015learning} to extract its frame-level features $\mathbf{V} = \{\mathbf{v}_i\}^n_{i=1} \in \mathbb{R}^{n \times D_v}$, where $n$ is the number of frames and $D_v$ is the feature dimension. 
For each query, we deploy the GloVe model~\cite{pennington2014glove} to obtain word-level embeddings $\mathbf{Q} = \{\mathbf{q}_i\}^m_{i=1}\in\mathbb{R}^{m \times D_q}$, where $m$ is the number of the words and $D_q$ is the feature dimension.
Then both video and query features are projected into the same latent space by two fully-connected layers to generate $\mathbf{V}^\prime$ and $\mathbf{Q}^\prime$ with the same dimension $D$, where $\mathbf{V}^\prime\in\mathbb{R}^{n \times D}$ and $\mathbf{Q}^\prime\in\mathbb{R}^{m \times D}$.
After that, we feed $\mathbf{V}^\prime, \mathbf{Q}^\prime$ into the modality-specific encoders $f_v(\cdot), f_q(\cdot)$ to generate the final visual representations $\widetilde{\mathbf{V}}$ and query embeddings $\widetilde{\mathbf{Q}}$.
That is, $\widetilde{\mathbf{V}}=f_v(\mathbf{V}^\prime), \widetilde{\mathbf{Q}} = f_q(\mathbf{Q}^\prime)$. 
Here, $f_v(\cdot), f_q(\cdot)$ share weights and consist of four convolution layers, followed by a multi-head attention layer \cite{vaswani2017attention}.

\noindent \textbf{Video-query interaction.}
We further apply a video-query attention mechanism to learn the cross-modal interactions, where we calculate the similarity score $\mathbf{S}\in\mathbb{R}^{n \times m}$ between video and query features, and use the SoftMax operation along the row and column to generate $\mathbf{S}_r$ and $\mathbf{S}_c$, respectively.
Next, we compute the video-to-query ($\mathcal{V}$) and query-to-video ($\mathcal{Q}$) attention contexts \cite{zhang2020span} as:
\begin{equation}
    \mathcal{V} = \mathbf{S}_r \cdot \widetilde{\mathbf{Q}} \in \mathbb{R}^{n \times D},\\
    \mathcal{Q} = \mathbf{S}_r \cdot \mathbf{S}_c^T \cdot \widetilde{\mathbf{V}} \in \mathbb{R}^{m \times D}.
\end{equation}
Then a single feed-forward layer $\mathtt{FFN}$ (composed of multiple linear layers) is applied to generate the interacted output features $\mathbf{V}^q \in\mathbb{R}^{n\times D}$ as:
\begin{equation}\label{eq:vq}
    \mathbf{V}^q = \mathtt{FFN}(\widetilde{\mathbf{V}};\mathcal{V}; \widetilde{\mathbf{V}}\odot\mathcal{V}; \widetilde{\mathbf{V}}\odot\mathcal{Q}),
\end{equation}
where $\odot$ denotes the Hadamard product.






\subsection{Frame-wise Matching Score Prediction}
In the weakly-supervised setting, we only have access to the knowledge of the matched video-query pair without corresponding detailed segment-level annotations.
In order to determine which frame is matched with the query semantics and how much the frame contributes to the final grounding, we introduce a score-based self-supervised branch to predict frame-wise matching scores and frame-wise matching weights for choosing the most possible foreground frames.
Specifically, we devise a frame score head $h_s(\cdot)$ and a frame weight head $h_w(\cdot)$ to predict the corresponding matching score $s_i$ and weight $w_i$ for each frame $i$, respectively. 
Here, both $h_s(\cdot), h_w(\cdot)$ are composed of three linear layers. 
The $S=\{s_i\}_{i=1}^n$ and $W=\{w_i\}_{i=1}^n$ are formulated as: 
\begin{equation}\label{eq:h_s_w}
    S = \mathtt{Sigmoid}(h_s(\mathbf{V}^q)); W = \mathtt{Softmax}(h_w(\mathbf{V}^q)). 
\end{equation}

Then, for the $k$-th video and $k$-th query in each batch, their final semantic matching score $\hat{s}_{k,k}$ is calculated as $\hat{s}_{k,k} = \sum_{i=1}^n s_i \cdot w_i$. 
In addition, we also estimate the similarity score (utilizing dot-product attention) between video features $\widetilde{\mathbf{V}}$ and query features $\widetilde{\mathbf{Q}}$ to measure their distance.
The overall score objective is defined as:
\begin{equation}\label{eq:score}
    \mathcal{L}_{\text{score}} = -\log\frac{\sum_{k=1}^K (\hat{s}_{k,k} + \widetilde{\mathbf{V}}_k \cdot \widetilde{\mathbf{Q}}_k)}{\sum_{k=1}^K\sum_{j=1}^K (\hat{s}_{k,j} + \widetilde{\mathbf{V}}_k \cdot \widetilde{\mathbf{Q}}_j)},
\end{equation}
where $\hat{s}_{k,j}$ represents the overall video score corresponding to the $j$-th query features and $k$-th video features at the same batch. $\hat{s}_{k,k}$ denotes the score of the matched video-query pair.
$K$ denotes the batch size.
In this way, we maximize the overall score of video and query features from correct pairs while minimizing the score of false pairs.
After getting the matching scores of all frames, we take them as pseudo labels to provide better supervisions for iteratively training the following contrastive learning module, and the learned discriminative features in turn further lead to more precise matching score prediction.

\subsection{Multi-scale Self-contrastive Learning}
In order to discriminate the frame-wise representations for more accurate prediction of matching scores, we propose a multi-scale self-contrastive learning paradigm with hard negative mining to capture more discriminative frame-wise representations in a coarse-to-fine manner.
Specifically, we iteratively mine hard negative samples that are close to positive samples in the representation space with a multi-step strategy.
In each step, we first choose the positive and negative frames according to their predicted frame-wise scores at frame-scale, and then consider one positive segment with the highest segment score while taking the other segments as negative samples at frame-scale. Then, we perform both frame-scale and segment-scale contrastive learning to learn more discriminative fine-grained frame-wise details. 
The updated frame-wise features in turn provide more accurate matching score to mine harder negative samples in the next step of learning.
By performing multi-scale self-contrastive learning with such an iterative strategy, our model is able to enforce more accurate segment grounding.
We will illustrate the details of both frame- and segment-scale negative mining of each step in the following.

\input{SECTIONS/30_Methodology/algo_main}

%



\input{SECTIONS/40_Experiments/exp_charades_activitynet}

\noindent{\textbf{Frame-scale.}}
In order to mine hard negative frames that are close to positive frames in the representation space, we iteratively assign a lower bound $b_l$ and a upper bound $b_u$ to select positive and negative frames.
The lower bound $b_l$ and the upper bound $b_u$ are defined from frame-wise scores as:
\begin{equation}\label{eq:bound}
    b_l = b_l^0 * \delta^{e-e_0}, \
    b_u = \frac{1}{n}\sum_{i=1}^n s_i,
\end{equation}
where $\delta$ is the increasing step and $e, e_0$ denote the current epoch and the updated cycle of epoch respectively. $b_l^0$ is the initial value of $b_l$. We set $b_l^0=e^{-8}$, $e_0=50, \delta=10$ during the training, that is, we increase $b_l$ exponentially by 10 every 50 epoch after the warm-up stage.

Accordingly, we consider frames with scores greater than $b_u$ as positive frames, and other frames with scores ranging from $b_l$ to $b_u$ as negative frames. The loss function of the frame-scale contrastive learning is defined as:
\begin{equation}\label{eq:frame}
    \mathcal{L}_{\text{fra}} = -\log\frac{\sum_{k=1}^K \widetilde{\mathbf{V}}_k \cdot \widetilde{\mathbf{Q}}_k \cdot \mathbf{p}_{k}^f}{\sum_{k=1}^K \widetilde{\mathbf{V}}_k \cdot \widetilde{\mathbf{Q}}_k \cdot\mathbf{n}_{k}^f},
\end{equation}
where $\mathbf{p}_k^f$ and $\mathbf{n}_k^f$ denote the binary index mask of positive and negative frames at batch index $k$, respectively. The entries of corresponding indices are 1 and others are 0.

\noindent{\textbf{Segment-scale.}}
In order to enforce more accurate segment grounding predictions, we locate the predicted segments $\{g_t\}_{t=1}^T$ with consecutive indices and calculate the segment score by averaging the scores of internal frame located in the segment.
Then we consider the segment with the highest segment score as the positive segment and other segments as negative samples.
The loss function of the segment-scale contrastive learning is formulated as:
\begin{equation}\label{eq:segment}
    \mathcal{L}_{\text{seg}} = -\log\frac{\sum_{k=1}^K \widetilde{\mathbf{V}}_k \cdot \widetilde{\mathbf{Q}}_k \cdot \mathbf{p}_{k}^g}{\sum_{k=1}^K \widetilde{\mathbf{V}}_k \cdot \widetilde{\mathbf{Q}}_k \cdot\mathbf{n}_{k}^g},
\end{equation}
where $\mathbf{p}_k^g$ and $\mathbf{n}_k^g$ denote the binary index mask of frames located in positive and negative segments at batch index $k$, separately. The entries of corresponding indices are 1 and the others are 0.

The overall objective of our model is minimized in an end-to-end manner and formulated as:
\begin{equation}\label{eq:all}
    \mathcal{L} =  \mathcal{L}_{\text{score}} + \lambda_{\text{fra}}\cdot\mathcal{L}_{\text{fra}} + \lambda_{\text{seg}}\cdot \mathcal{L}_{\text{seg}},
\end{equation}
where $\lambda_{\text{fra}}$ and $\lambda_{\text{seg}}$ denote the weighting hyper-parameters of the frame loss and segment loss, respectively.
In the experiments, we set $\lambda_{\text{fra}}=10$ and $\lambda_{\text{seg}}=5$. 

The overall algorithm of our training approach is summarized in Algorithm~\ref{alg:main}, where we utilize an iterative strategy to gradually mine the hard negative samples.
In order to enforce the model predict accurate positive samples with higher confidence,
we first warm-up our model in the first 50 epochs without the procedure of hard negative sampling. Then, we iteratively mine the hard negative samples with high similarity to the positive ones.

\subsection{Segment Localization}

At the inference time, we select the segment with the highest segment score as the final prediction. 
Specifically, we extract all possible segments by choosing consecutive frames with scores higher than the upper bound $b_u$, and calculate the segment score by averaging the scores of internal frames located in the segment.
Then we take the segment with the highest segment score as the final output.

%% file: SECTIONS/30_Methodology/algo_main.tex
\begin{algorithm}[t]
   \caption{Multi-scale contrastive learning algorithm}
   \label{alg:main}
\begin{algorithmic}[1]
   \Statex {\bfseries Input:} video and query features $\mathbf{V}$ and $\mathbf{Q}$,
   iteration number $L$ 
   \State Initialize the parameters $f_v(\cdot), f_q(\cdot), h_s(\cdot), h_w(\cdot), b_l, b_u$
   
   \State Warm-up our model for 50 epochs without hard negative sampling
   
   \For{iteration $l\gets 1$ \text{to} $L$}
   \State Encode features $\mathbf{V}, \mathbf{Q}$ and calculate $\mathbf{V}^q$ as in Eq.~\ref{eq:vq}
   \State Predict frame scores and weights as in Eq.~\ref{eq:h_s_w}
   \State Calculate the score loss as in Eq.~\ref{eq:score}
   \State Update $b_l, b_u$ in Eq.~\ref{eq:bound}
   \State Calculate frame and segment losses in Eq.~\ref{eq:frame} and ~\ref{eq:segment}
   
\State Compute the total loss in Eq.~\ref{eq:all}
   \State Update the parameters of $f_v(\cdot), f_q(\cdot), h_s(\cdot), h_w(\cdot)$
    \EndFor
    \Statex {\bfseries Output:} $f_v(\cdot), f_q(\cdot), h_s(\cdot), h_w(\cdot)$
\end{algorithmic}
\end{algorithm}

%% file: SECTIONS/40_Experiments/exp_charades_activitynet.tex
\begin{table*}[!htb]
	\centering
	\scalebox{0.8}{
		\begin{tabular}{|l|ccc|ccc|ccc|ccc|}
		    \hline
		    \multicolumn{1}{|c|}{\multirow{3}{*}{Method}} &\multicolumn{6}{c|}{Charades-STA} &\multicolumn{6}{c|}{ActivityNet-Caption}
		    \\ \cline{2-13}
		    \multicolumn{1}{|c|}{} &\multicolumn{3}{c|}{R@1} & \multicolumn{3}{c|}{R@5} &\multicolumn{3}{c|}{R@1} & \multicolumn{3}{c|}{R@5}        \\
\multicolumn{1}{|c|}{}  & IoU=0.3 & IoU=0.5 & IoU=0.7 & IoU=0.3 & IoU=0.5 & IoU=0.7 & IoU=0.1 & IoU=0.3 & IoU=0.5 & IoU=0.1 & IoU=0.3 & IoU=0.5 \\
			\hline
			\hline
			TGA & 32.14 & 19.94 & 8.84 & 86.58 & 65.52 & 33.51 & - & - & - &  - & - & - \\
			CTF & 39.80 & 27.30 & 12.90 & - & - & - & 74.20	& 44.30 & 23.60 & - & - & -\\
			ReLoCLNet & - & - & - &  - & - & -  & - & 42.65 & 28.54 & - & - & -\\
			SCN  & 42.96 & 23.58 & 9.97 & 95.56 & 71.80 & 38.87 & 71.48 & 47.23 & 29.22 & - & 71.45 & 55.69 \\
			MARN & - & 31.94 & 14.81 & -& 70.00 & 	37.40 & 	- & 47.01 & 29.95 & - & 72.02 & 57.49 \\
			RTBPN & \textbf{60.04} & 32.36 & 13.24 & 97.48 & 71.85 & 41.18 & 73.73 & 49.77 & 29.63 & 93.89 & 79.89 & 60.56 \\
			VGN+CCL & -	& 33.21 & 15.68	& - & 73.50 & 41.87 & -	& 50.12 & 31.07 & - & 77.36 & 61.29 \\ \hline \hline
			\textbf{Ours} & 58.92 & \textbf{43.15} & \textbf{23.49} & \textbf{98.02} & \textbf{81.23} & \textbf{48.45} & \textbf{75.61} & \textbf{55.05} & \textbf{38.23} & \textbf{95.26} & \textbf{82.72} & \textbf{68.05} \\
            
			\hline
			\end{tabular}}
	\caption{Comparison results on the Charades-STA and ActivityNet-Caption datasets.}
	\label{tab: exp_charades_activitynet}
	\renewcommand\tabcolsep{4.0pt}
	\vspace{-1em}
\end{table*}

%% file: SECTIONS/40_Experiments/manuscript.tex
\section{Experiments}

\subsection{Datasets and Evaluation Metrics}

\noindent \textbf{Charades-STA.}
The Charades-STA dataset \cite{gao2017tall} is built based on the Charades \cite{sigurdsson2016hollywood} dataset, which contains 6,672 videos of indoor activities and involves 16,128 query-video pairs. There are 12,408 pairs used for training and 3,720 used for testing. The average duration of each video is 29.76 seconds. Each video has 2.4 annotated moments and each annotated moment lasts for 8 seconds on average.

\noindent \textbf{ActivityNet-Caption.}
The ActivityNet-Caption dataset \cite{krishna2017dense} contains 20,000 videos with 100,000 queries, where 37,421 query-video pairs are used for training and 34,536 are used for testing. The average duration of the videos is 1 minute and 50 seconds. On average, each video in ActivityNet-Caption has 3.65 annotated moments and each annotated moment lasts for 36 seconds.

\noindent \textbf{Evaluation metrics.}
Following previous works, we adopt the metrics ``R@n, IoU=m" to evaluate our model, where ``R@n, IoU=m" presents the proportion of the top n moment candidates with IoU larger than m. Specifically, we set n as 1, 5 and set m as 0.3, 0.5, 0.7 in Charades-STA dataset and 0.1, 0.3, 0.5 in ActivityNet-Caption dataset.

\subsection{Experimental Settings}

To make a fair comparison with previous methods like ~\cite{zhang2020counterfactual}, we extract video features from the pre-trained C3D network~\cite{du2015learning} and query features from the 300-d Glove embedding~\cite{pennington2014glove}.
We train the model for 200 epochs with the batch size of 16. 
We use a warm-up training without hard negative sampling for 50 epochs.
The dimension of encoded features is set to 512.
Our model is optimized by Adam~\cite{kingma2014adam} with an initial learning rate of 0.01 and linear decay of learning rate.
All experiments are conducted on single NVIDIA GeForce RTX 3090 GPU.

\subsection{Comparison with State-of-the-art Methods}

\noindent{\textbf{Charades-STA.}}
We compare our method against the current state-of-the-art methods under weakly-supervised settings on Charades-STA dataset in Table~\ref{tab: exp_charades_activitynet}.
As can be seen, we achieve the best performance over all baselines.
Particularly, our MSCL outperforms VGN+CCL~\cite{zhang2020counterfactual}, the current state-of-the-art method, by a large margin 9.94\%, 7.81\% and 7.73\%, 6.58\% in terms of R@1,IoU=0.5, 0.7 and R@5,IoU=0.5, 0.7, respectively.
This indeed shows the superiority of our multi-scale hard negative sampling strategy in weakly-supervised video grounding.

\noindent{\textbf{ActivityNet-Caption.}}
Table~\ref{tab: exp_charades_activitynet} also reports the comparison results with the state-of-the-art methods on ActivityNet-Caption dataset.
We can observe that our MSCL achieves superior performance against all previous works. 
This further demonstrates the advantage of the multi-scale hard negative sampling on  distinguishing fine-grained frame-wise details and enforcing more accurate segment grounding.

\input{SECTIONS/50_Ablation_Study/ab_hnm}

%% file: SECTIONS/50_Ablation_Study/ab_hnm.tex
\begin{table}[t]
	
	\renewcommand\tabcolsep{4.0pt}
	\centering
	\scalebox{0.85}{
		\begin{tabular}{|ccc|ccc|}
		    \hline
	    \multicolumn{1}{|c}{\multirow{2}{*}{$\mathcal{L}_\text{score}$}} & \multicolumn{1}{c}{\multirow{2}{*}{$\mathcal{L}_\text{fra}$}} & \multicolumn{1}{c|}{\multirow{2}{*}{$\mathcal{L}_\text{seg}$}} &\multicolumn{3}{c|}{R@1}         \\
\multicolumn{1}{|c}{}  & \multicolumn{1}{c}{}  & \multicolumn{1}{c|}{}  & IoU=0.3 & IoU=0.5 & IoU=0.7  \\
			\hline
			\hline
			\xmark & \xmark & \xmark & 44.19$\pm$0.25 & 19.95$\pm$0.21 & 8.60$\pm$0.18 \\ 
			\cmark & \xmark & \xmark & 50.91$\pm$0.19 & 21.13$\pm$0.18 & 9.54$\pm$0.16 \\ 
            \cmark & \cmark & \xmark & 55.22$\pm$0.12 & 29.73$\pm$0.09 & 14.52$\pm$0.08 \\ 
            \cmark & \xmark & \cmark & 51.94$\pm$0.16 & 33.66$\pm$0.11 & 16.77$\pm$0.09 \\ 
			\cmark & \cmark & \cmark & \textbf{58.65}$\pm$\textbf{0.08} & \textbf{42.92}$\pm$\textbf{0.05} & \textbf{23.25}$\pm$\textbf{0.03} \\

			\hline
			\end{tabular}}
	\caption{Ablation study for the effect of each module.}
	\label{tab: ab_hnm}
	\vspace{-0.5em}
\end{table}

%% file: SECTIONS/50_Ablation_Study/manuscript.tex
\subsection{Ablation Study}

In this part, we conduct extensive ablation studies on each module in our MSCL including frame-wise matching score prediction and frame-/segment-scale hard negative sampling, the effect of batch size, and the hyper-parameters. 
Unless specified, we perform all ablation studies on the Charades-STA benchmark.

\noindent{\textbf{Effect of each module.}}
In order to understand how each module in our MSCL affects the final performance, we explore the effect of each proposed loss as shown in Table~\ref{tab: ab_hnm}.
Our model with frame-wise matching score prediction outperforms the baseline by 6.72\%, 1.18\%, and 0.94\% in terms of three criteria, which shows the effectiveness of this module.
Introducing frame-scale and segment-scale hard negative sampling separately boosts the performance of our model with frame-wise matching score prediction only.
Furthermore, by adding frame-scale and segment-scale hard negative sampling together, we observe the highest increasing range of 7.74\%, 21.79\%, and 13.71\%.  
This demonstrates the superiority of our multi-scale hard negative sampling over baselines.

\input{SECTIONS/50_Ablation_Study/ab_loss}

\input{SECTIONS/50_Ablation_Study/ab_bs}

\begin{figure*}[!htb]
	    \centering
\includegraphics[width=\linewidth]{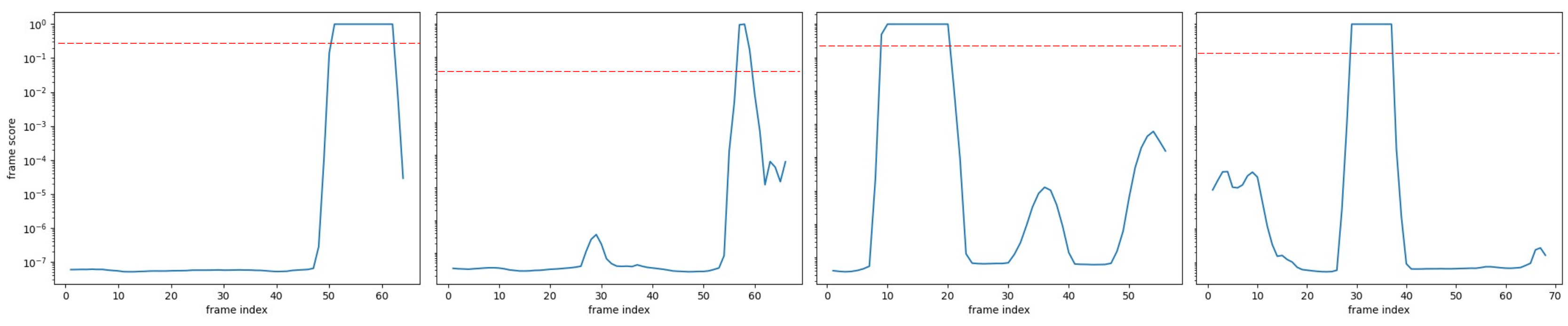}
   \caption{Visualization of frame score curves used for segment localization. Red dotted lines denote the upper bound $b_u$. The ground-truth of frame indices are 49-60, 53-62, 6-23, and 26-37.}
	\label{fig: vis_score}
\end{figure*}

\noindent{\textbf{Analysis of frame and segment loss.}}
Furthermore, we conduct extensive experiments to explore the weighting hyper-parameters of frame and segment loss used in multi-scale hard negative sampling in Table~\ref{tab: ab_loss}. 
Specifically, we take the value of $\lambda_{fra}$ and $\lambda_{seg}$ from (1, 5, 10) for different control settings.
With the increase of $\lambda_{fra}$, the performance of our model decreases since we fail to consider the more global information of segments in the whole video.
However, with the increase of $\lambda_{seg}$, more global information of segments is introduced such that the results of our MSCL are improved.
As can be seen, our MSCL reaches the best performance when $\lambda_{fra}$=10 and $\lambda_{seg}$=5, which implies the importance of balancing the weight of the frame-wise and segment-wise hard negative sampling during the training.

\noindent{\textbf{Robustness to the batch size.}}
In this part, we analyze the effect of the batch size on the final performance of our MSCL, as shown in Table~\ref{tab: ab_bs}, where we set the batch size as 4, 8, 16, 32, 48.
From Table~\ref{tab: ab_bs}, we observe that our MSCL with the batch size of 16 achieves the best results in terms of all metrics. 
Meanwhile, the performance of our model does not change too much with the altering of the batch size.
This further validates the robustness of our MSCL to the choice of the batch size.
In other words, we do not need a large batch size that is desired in previous contrastive learning based methods~\cite{zhang2020counterfactual,zhang2021video} for weakly-supervised video grounding.

\begin{figure*}[!htb]
\setlength{\abovecaptionskip}{0em}
\setlength{\belowcaptionskip}{-1em}
	    \centering
		\includegraphics[width=0.95\linewidth]{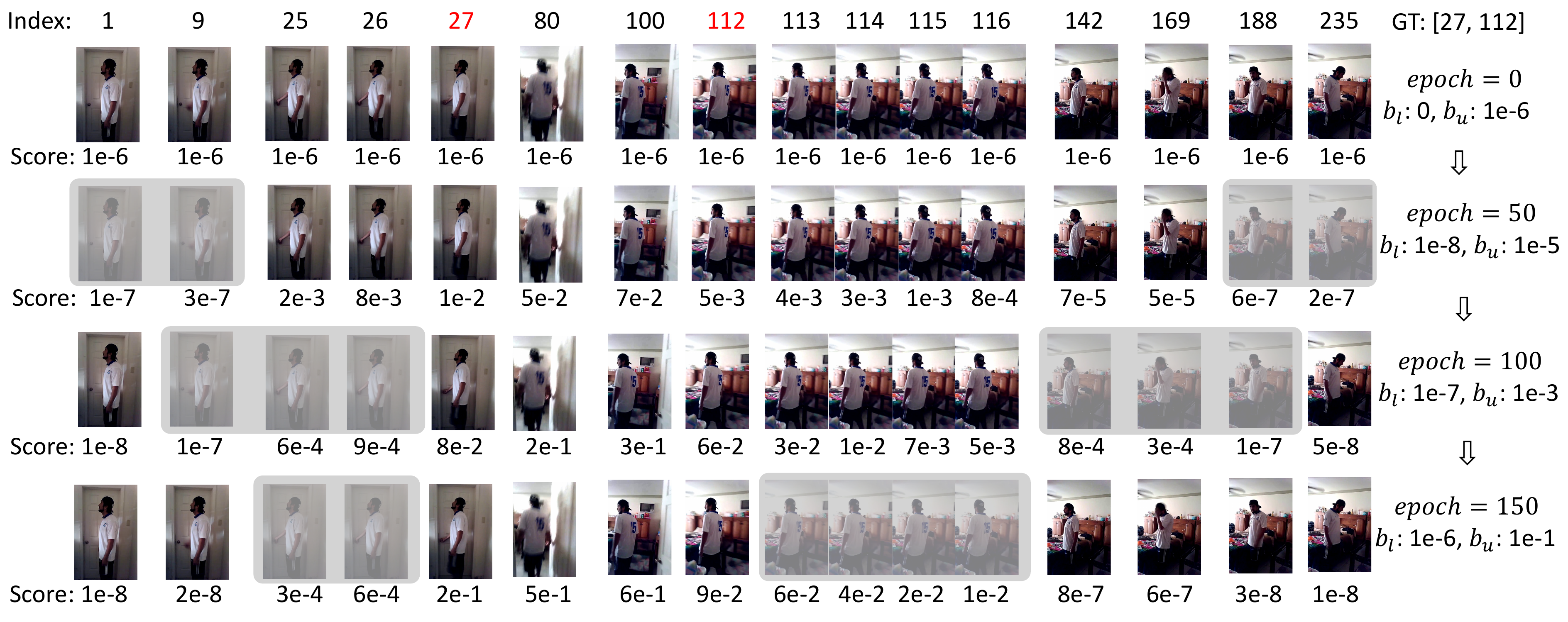}
   \caption{An illustration of the dynamic process of hard negative sampling (Gray Shadow) at epoch=0, 50, 100, 150. GT denotes the ground-truth (red indexes denote the segment boundaries of ground-truth). It shows that our iterative mining strategy can mine harder negative samples with the step goes on, leading to more discriminative frame-wise representation learning.}
	\label{fig: vis_hnm}
\end{figure*}

\input{SECTIONS/40_Experiments/exp_vis}

\subsection{Visualization Results}

In this section, we provide more detailed visualization results on how our MSCL predicts more accurate segment grounding results given frame score curves.
Qualitative examples of hard negative sampling and foreground frame predictions on two benchmarks are visualized to validate the superiority of our MSCL.

\noindent{\textbf{Frame-wise matching scores.}}
In order to better understand the effectiveness of the segment localization in our MSCL, we plot the frame-wise score curves with respect to the frame index among the positive frames in Figure~\ref{fig: vis_score}.
As can be seen, many hard negative samples with high frame scores appeared in the training process. 
With the multi-scale hard negative sampling, our MSCL predicts the segment with the highest score greater than the upper bound $b_u$ as the final output, which matches the target segment. 
This shows the importance of the segment localization in more accurate segment grounding.

\noindent{\textbf{Hard Negative Mining.}}
In Figure~\ref{fig: vis_hnm}, we also visualize the hard negative samples at epoch=0, 50, 100, 150.
We observe the hard negative samples with high scores are progressively closer to the positive frames, which validates the effectiveness of our multi-scale hard negative sampling.

\noindent{\textbf{Qualitative Results.}}
To validate the superiority of our MSCL in a qualitative manner, we visualize qualitative examples of Charades-STA and ActivityNet-Caption benchmarks in Figure~\ref{fig: vis_charades_activitynet}.
By comparison, our MSCL achieves better performance than SCN~\cite{lin2020weakly} and VGN+CCL~\cite{zhang2020counterfactual}.
Particularly, we achieve more accurate results on the boundary of the ground-truth segment due to the effectiveness of our multi-scale hard negative sampling strategy.

%% file: SECTIONS/50_Ablation_Study/ab_loss.tex
\begin{table}[t]
	\renewcommand\tabcolsep{4.0pt}
	\centering
	\scalebox{0.9}{
		\begin{tabular}{|cc|ccc|}
		    \hline	\multicolumn{1}{|c}{\multirow{2}{*}{$\lambda_\text{fra}$}} & \multicolumn{1}{c|}{\multirow{2}{*}{$\lambda_\text{seg}$}} & \multicolumn{3}{c|}{R@1}         \\
\multicolumn{1}{|c}{}  & \multicolumn{1}{c|}{}  & IoU=0.3 & IoU=0.5 & IoU=0.7  \\
			\hline
			\hline
			1 & 1  & 58.65$\pm$0.08 & 42.92$\pm$0.05 & 23.25$\pm$0.03 \\
		    5 & 1  & 58.01$\pm$0.07 & 42.58$\pm$0.05 & 22.75$\pm$0.03 \\
            10 & 1 & 58.22$\pm$0.07 & 42.75$\pm$0.04 & 22.96$\pm$0.02 \\
            1 & 5  & 58.85$\pm$0.05 & 43.07$\pm$0.03 & 23.36$\pm$0.01 \\
            1 & 10 & 58.52$\pm$0.07 & 42.87$\pm$0.04 & 23.18$\pm$0.02 \\
			10 & 5 & \textbf{58.92}$\pm$\textbf{0.04} & \textbf{43.15}$\pm$\textbf{0.02} & \textbf{23.49}$\pm$\textbf{0.01} \\
			\hline
			\end{tabular}}
	\caption{Ablation study for $\lambda_\text{fra}$ and $\lambda_\text{seg}$.}
	\vspace{-1em}
	\label{tab: ab_loss}
\end{table}

%% file: SECTIONS/50_Ablation_Study/ab_bs.tex
\begin{table*}[!tbh]
	\renewcommand\tabcolsep{4.0pt}
	\centering
	\scalebox{0.85}{
		\begin{tabular}{|c|ccc|ccc|}
		    \hline	\multicolumn{1}{|c|}{\multirow{2}{*}{Batch Size}} &\multicolumn{3}{c|}{R@1} & \multicolumn{3}{c|}{R@5}         \\
\multicolumn{1}{|c|}{}  & IoU=0.3 & IoU=0.5 & IoU=0.7 & IoU=0.3 & IoU=0.5 & IoU=0.7 \\
			\hline
			\hline
			4  & 58.53$\pm$0.08 & 42.81$\pm$0.06 & 23.19$\pm$0.04 & 97.88$\pm$0.05 & 80.97$\pm$0.04 & 48.12$\pm$0.04 \\
			8 & 58.81$\pm$0.05 & 43.03$\pm$0.03 & 23.38$\pm$0.02 & 97.96$\pm$0.04 & 81.14$\pm$0.04 & 48.36$\pm$0.03 \\
			16  & \textbf{58.92}$\pm$\textbf{0.03} & \textbf{43.15}$\pm$\textbf{0.01} & \textbf{23.49}$\pm$\textbf{0.01} & \textbf{98.02}$\pm$\textbf{0.02} & \textbf{81.23}$\pm$\textbf{0.02} & \textbf{48.45}$\pm$\textbf{0.01} \\
			32 & 58.91$\pm$0.02 & 43.12$\pm$0.01 & 23.45$\pm$0.01 & 97.98$\pm$0.01 & 81.19$\pm$0.01 & 48.42$\pm$0.01 \\
			48 & 58.85$\pm$0.02 & 43.09$\pm$0.02 & 23.41$\pm$0.01 & 97.92$\pm$0.02 & 81.13$\pm$0.02 & 48.36$\pm$0.02 \\
            
			\hline
			\end{tabular}}
	\caption{Exploration study on the effect of batch size.}
	\label{tab: ab_bs}
	\vspace{-1em}
\end{table*}

%% file: SECTIONS/40_Experiments/exp_vis.tex
\begin{figure*}[!htb]
\setlength{\abovecaptionskip}{0em}
\setlength{\belowcaptionskip}{-0.5em}
	    \centering
		\includegraphics[width=0.95\linewidth]{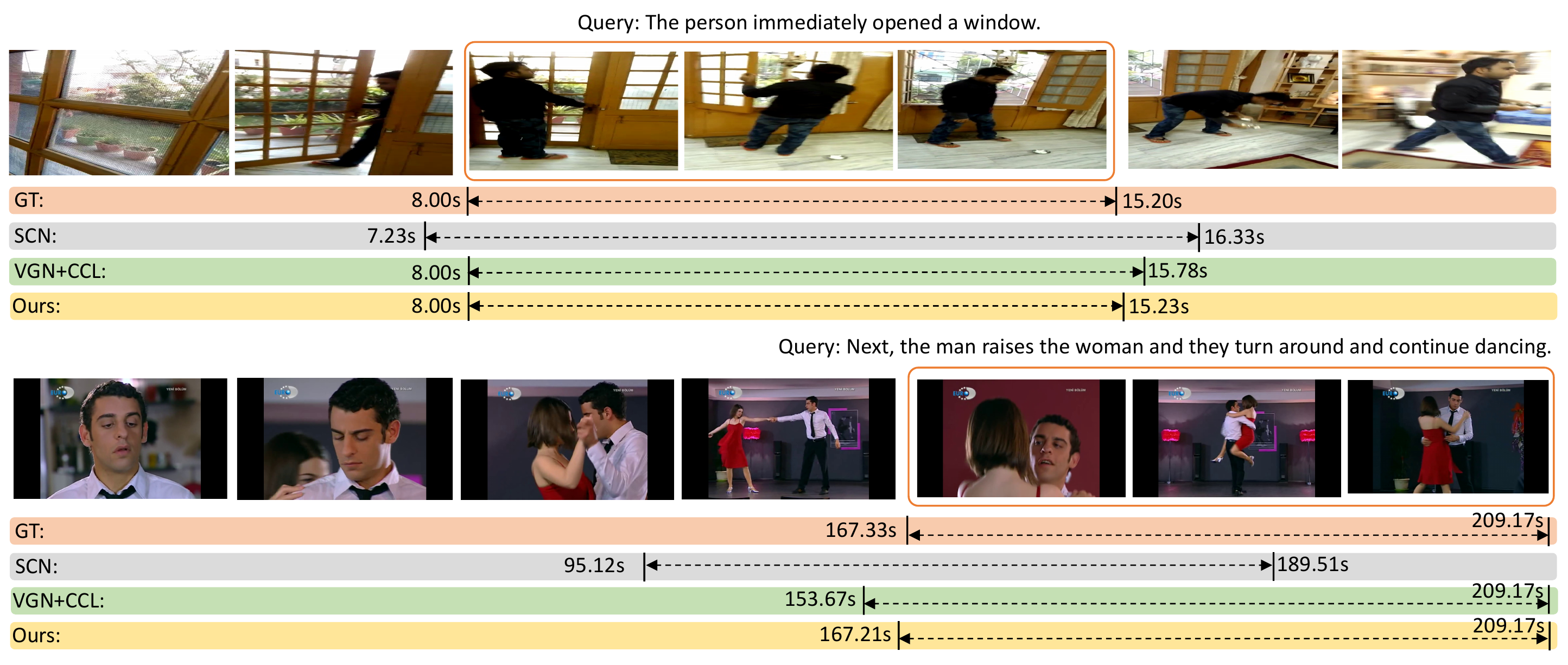}
   \caption{Visualization of examples on Charades-STA and ActivityNet-Caption  benchmarks. GT denotes the ground-truth.}
	\label{fig: vis_charades_activitynet}
	\vspace{-0.5em}
\end{figure*}

%% file: SECTIONS/60_Conclusion/manuscript.tex
\section{Conclusion}
In this work, we propose a novel multi-scale self-contrastive learning model for weakly-supervised query-based video grounding.
Instead of utilizing redundant segment proposal for semantic matching, we predict frame-wise scores and weights for matching fine-grained frame-wise features with query semantics. 
In order to learn more discriminative frame-wise representations for predicting accurate frame-wise scores, we further introduce a multi-scale self-contrastive learning with multi-step hard negative mining strategy to progressively discriminate hard negative samples that are close to positive samples in the representation space.
This iterative approach captures fine-grained frame-scale details as well as segment-scale semantics for distinguishing frames with high repeatability and similarity within the entire video.
Experimental results show that our proposed model outperforms state-of-the-art methods on two challenging benchmarks.